\documentclass[conference]{IEEEtran}
\IEEEoverridecommandlockouts
\usepackage{amsmath,amssymb,amsfonts}
\usepackage{algorithmic}
\usepackage{graphicx}
\usepackage{textcomp}
\usepackage{xcolor}
\def\BibTeX{{\rm B\kern-.05em{\sc i\kern-.025em b}\kern-.08em
    T\kern-.1667em\lower.7ex\hbox{E}\kern-.125emX}}

\usepackage{hyperref}
\usepackage[style=numeric, sorting=none, maxnames=6]{biblatex}
\usepackage{csquotes}
\usepackage[utf8]{inputenx}

\usepackage{listings}
\usepackage{xcolor}
\begin{document}

\title{Tackling fake images in cybersecurity\\- Interpretation of a StyleGAN and lifting its black-box}

\author{\IEEEauthorblockN{Julia Laubmann; Johannes Reschke}
\IEEEauthorblockA{\textit{Faculty of Electrical Engineering and Information Technology} \\
\textit{Ostbayrische Technische Hochschule Regensburg}\\
Seybothstraße 2, 93053 Regensburg \\
julia.laubmann@st.oth-regensburg.de; johannes.reschke@oth-regensburg.de}}

\maketitle

\begin{abstract}
	In today's digital age, concerns about the dangers of AI-generated images are increasingly common. One powerful tool in this domain is StyleGAN (style-based generative adversarial networks), a generative adversarial network capable of producing highly realistic synthetic faces. To gain a deeper understanding of how such a model operates, this work focuses on analyzing the inner workings of StyleGAN’s generator component. Key architectural elements and techniques, such as the Equalized Learning Rate, are explored in detail to shed light on the model’s behavior. A StyleGAN model is trained using the PyTorch framework, enabling direct inspection of its learned weights. Through pruning, it is revealed that a significant number of these weights can be removed without drastically affecting the output, leading to reduced computational requirements. Moreover, the role of the latent vector - which heavily influences the appearance of the generated faces - is closely examined. Global alterations to this vector primarily affect aspects like color tones, while targeted changes to individual dimensions allow for precise manipulation of specific facial features. This ability to fine-tune visual traits is not only of academic interest but also highlights a serious ethical concern: the potential misuse of such technology. Malicious actors could exploit this capability to fabricate convincing fake identities, posing significant risks in the context of digital deception and cybercrime.
\end{abstract}


\section{Introduction}
\label{sec: Introduction}
In today's modern age, models to generate human faces are based on StyleGANs, which have been trained with the FFHQ (Flickr-Faces-High-Quality) image dataset \cite{karras_nvlabsffhq-dataset_2019}. This type of GAN enables a style transfer so that characteristics, such as facial features or accessories, of an existing image can be transferred to a new one. The generation process starts with a low resolution, increasing it iteratively, whereby resolutions of up to 1024\,x\,1024 pixels can be achieved with this method.

The German Federal Office for Information Security (BSI) draws attention to the danger that content, especially images, in media can be falsified using such deepfake methods. Faces, even in videos, can be exchanged, gestures and facial expressions can be changed and pseudo-identities can be created. As a result, there is a risk that biometric systems can be bypassed or that people's reputations can be damaged due to fake news. Spear fishing also poses a major danger of fraudsters using this technology to obtain confidential, sensitive data, documents and information as well as financial resources, which can have drastic consequences in all sectors and institutions. \cite{bundesamt_fur_sicherheit_in_der_informationstechnik_deepfakes_2022}

\begin{figure}[htbp]
	\centering
	\includegraphics[width=0.45\textwidth]{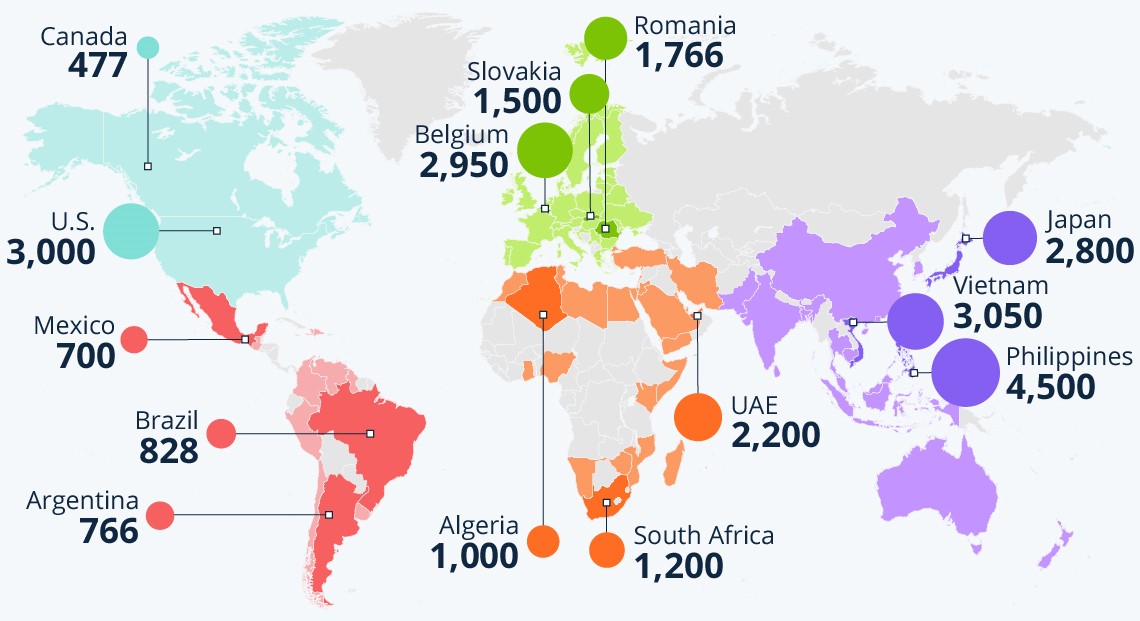}
	\caption{Countries per region with biggest increases in deepfake-specific fraud cases from 2022 to 2023 (in \%) \cite{zandt_how_2024}}
	\label{fig: Introduction_Growth_of_AI-Powered_Fraud2}
\end{figure}

As \autoref{fig: Introduction_Growth_of_AI-Powered_Fraud2} shows, AI-powered fraud cases have recently increased around the world. For this reason, the generated synthetic images and the structure of the StyleGAN are interpreted, which means an investigation and evaluation of the parameters used. In deep neural networks, necessary information can be obtained from the weights and activation functions, which is why these are analyzed in this contribution using various methods. The aim is to recognize characteristics relating to the units of the generative deep learning (DL) model in order to identify counterfeits. To achieve this, the black-box, which generates artificial images from real images, is transformed into a white-box using methods for interpreting the parameters it contains. \cite{zandt_how_2024, achille_where_2019}

\section{Basics of StyleGANs}
\label{sec: Basics of StyleGANs}
To appropriately interpret the network, it is essential to first examine the structure and functionality of its individual components in detail. For this purpose, a general overview of the GAN \cite{goodfellow_generative_2014}, developed by Ian J. Goodfellow and his team in 2014, will be provided. This will lay the foundation for identifying the unique features of StyleGAN \cite{karras_style-based_2018}, which was introduced five years later by researchers at NVIDIA.

\begin{figure}[htbp]
	\centering
	\includegraphics[width=0.45\textwidth]{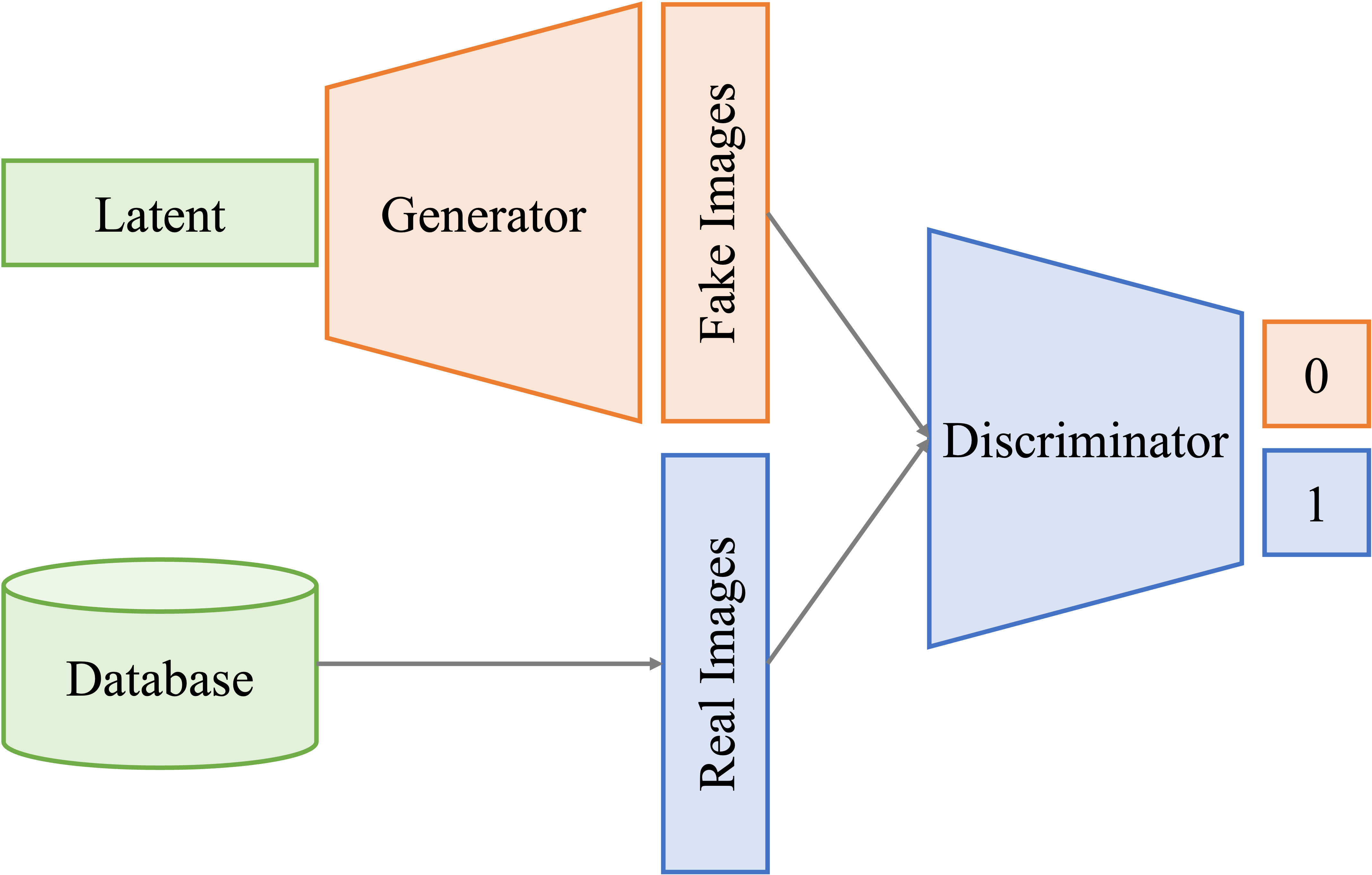}
	\caption{Basic structure of a GAN consisting of two networks for generating images (adapted from \cite{wenzel_generative_2023})}
	\label{fig: GAN_setup}
\end{figure}

The fundamental setup of a GAN is illustrated in \autoref{fig: GAN_setup}. From this, it can be deduced that such a model essentially consists of two networks: a generator and a discriminator. Its tasks include the generation of images, texts and music. When generating images, both counterparts are convolutional neural networks (CNNs). In general, the discriminator is trained to differentiate between real and generated solutions, while the generator focuses on producing the most realistic results possible. This creates an adversarial relationship between the two units based on their differing objectives. \cite{wenzel_generative_2023}

In the following, an explanation is given of how the generator and the discriminator for image generation work.

\subsection{Functioning of the generator}
\label{subsec: Functioning of the generator}
\begin{figure}[htbp]
	\centering
	\includegraphics[width=0.45\textwidth]{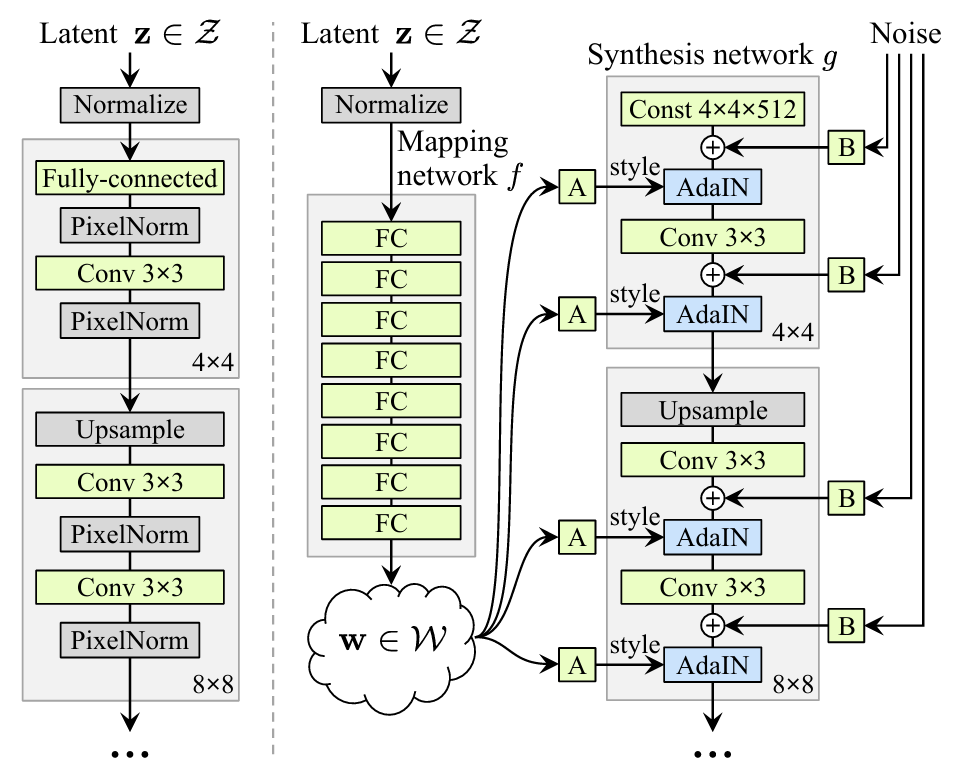}
	\caption{Comparison of the structure and functionality of a traditional and a style-based generator using the latent code $ \mathbf{z} \in \mathcal{Z} $ \cite{karras_style-based_2018}}
	\label{fig: traditional_vs_style-based_generator}
\end{figure}

The style-based generator differs from the traditional architecture in the way it works and its structure: Instead of using the latent code $ \mathbf{z} \in \mathcal{Z} $ directly as input for the generator, it is converted into another latent representation $ \mathbf{w} \in \mathcal{W} $ by an 8-layer MLP (multilayer perceptron) mapping network, as shown in \autoref{fig: traditional_vs_style-based_generator}. \cite{karras_style-based_2018}

The transformations "A" are transferred to the adaptive instance normalization (AdaIN) units at each convolution layer of the 18-layer synthesis network $ g $, which starts with a low resolution of $ 4 \times 4 $ and grows progressively to $ 1024 \times 1024 $. This serves to influence the style of each layer. The AdaIN operation is defined mathematically as follows:

\begin{equation}
	\text{AdaIN}(\mathbf{x}_i, \mathbf{y}) = \mathbf{y}_{s, i} \frac{\mathbf{x}_i - \mu(\mathbf{x}_i)}{\sigma(\mathbf{x}_i)} + \mathbf{y}_{b, i}
	\label{eq: AdaIN}
\end{equation}

Each feature map $ \mathbf{x}_i $ is normalized using its mean $ \mu(\mathbf{x}_i) $ and standard deviation $ \sigma(\mathbf{x}_i) $. The styles $ \mathbf{y} = (\mathbf{y}_s, \mathbf{y}_b) $ are then used to control the AdaIN operation: Scaling of the input is performed using $ \mathbf{y}_s $ and biasing using $ \mathbf{y}_b $. \cite{karras_style-based_2018}

To enable the generation of fine stochastic details, such as freckles and strands of hair, the generator receives explicit noise inputs. Each layer of the synthesis network is provided with a separate image containing uncorrelated Gaussian noise. This noise is scaled using learned per-feature coefficients and then added to the output of the corresponding convolutional layer. \cite{karras_style-based_2018}

The generator also allows style mixing by using several latent codes $ \mathbf{z}_1 $, $ \mathbf{z}_2 $ and the corresponding representations $ \mathbf{w}_1 $, $ \mathbf{w}_2 $. This possibility enables controlled feature transfer between generated images by selectively copying subsets of latent codes. Coarse-level latent codes influence global attributes such as pose, face shape, and hairstyle. Middle-level codes affect finer details like facial features and eye state, while fine-level codes primarily determine colors and textures. By blending latent codes from different sources, the generator creates hybrid images with a structured combination of characteristics. \cite{karras_style-based_2018}

\subsection{Tasks and application of the discriminator}
\label{subsec: Tasks and application of the discriminator}
As this work focuses on the creation of images, the functionality and benefits of the discriminator for the generator will only be briefly discussed.

The generator and the discriminator are opponents. This fact is mathematically formulated in \autoref{eq: MinMaxGD}. For the sake of simplicity and ease of understanding, the GAN here is a traditional GAN, as shown in \autoref{fig: GAN_setup}.

\begin{equation}
	\begin{aligned}
		\min_G \max_D V(D, G) &= \mathbb{E}_{\mathbf{x} \sim p_{\text{data}}(\mathbf{x})} [\log D(\mathbf{x})] \\
		&\quad + \mathbb{E}_{\mathbf{z} \sim p_{\mathbf{z}}(\mathbf{z})} [\log (1 - D(G(\mathbf{z})))]
	\end{aligned}
	\label{eq: MinMaxGD}
\end{equation}

In this adversarial framework, the generator $ G $ learns to map latent noise $ \mathbf{z} $ to data space, with a prior distribution $ p_{\mathbf{z}}(\mathbf{z}) $ on $ \mathbf{z} $. The discriminator $ D $, another multilayer perceptron, outputs the probability that an input $ \mathbf{x} $ is real or generated. The discriminator is trained to maximize the correct classification of real and generated samples, while the Generator is trained to minimize $ \log(1 - D(G(\mathbf{z}))) $, i.e., to fool the discriminator into classifying fake data as real. \cite{goodfellow_generative_2014}

In \autoref{sec: Interpretation of the StyleGAN}, the discriminator can be used to check whether changes in the generator parameters have a major influence on the quality of the generated images.

\subsection{The Equalized Learning Rate}
\label{subsec: The Equalized Learning Rate}
Instead of using careful weight initialization, a trivial $\mathcal{N}(0, 1)$ initialization is applied, and the weights are explicitly scaled at runtime. Specifically, the weights are defined as in \autoref{eq: Equal LR}.

\begin{equation}
	\hat{w}_i = \frac{w_i}{c}
	\label{eq: Equal LR}
\end{equation}

Here, $ w_i $ are the weights and $ c $ is the per-layer normalization constant from He's initializer (see \cite{he_delving_2015}). The advantage of applying this scaling dynamically, rather than during initialization, is subtle and relates to the scale-invariance of commonly used adaptive stochastic gradient descent methods such as RMSProp and Adam.

These methods normalize gradient updates by their estimated standard deviation, making the updates independent of the parameter scale. If certain parameters have a larger dynamic range than others, their updates may be slower. This is a situation that modern initializers can introduce, potentially causing the learning rate to be simultaneously too large and too small.

Dynamic scaling ensures consistent dynamic range and learning speed across all weights. \cite{karras_progressive_2018}

\subsection{Pruning for the reduction of computing power}
\label{subsec: Pruning for the reduction of computing power}
One method being investigated in the project is pruning, or more precisely magnitude pruning. Magnitude pruning is a widely used technique for introducing sparsity in neural networks. It works by identifying and setting to zero the weights with the smallest magnitudes - specifically those below a certain threshold. This threshold can be applied either within individual layers or across the entire network. Despite its simplicity, magnitude pruning is highly effective at producing sparse models and is considered a strong baseline approach for neural network sparsification. \cite{sun_simple_2024}

\section{Description of the datasets used}
\label{sec: Description of the datasets used}
The FFHQ dataset from NVIDIA is used in the first experiment. It contains 70,000 images of authentic faces with a resolution of $ 1024 \times 1024 $ in PNG format. These were obtained from users of the “Flickr” platform. The age and ethnicity of the people depicted and the background of the images vary. In addition, various accessories such as glasses or hats are worn, resulting in a high degree of diversity between the images. \cite{karras_style-based_2018, karras_analyzing_2019}

The network used in the following is also based on the StyleGAN and works with the MNIST dataset. This dataset contains 60,000 training images and 10,000 test images with $ 28 \times 28 $ pixels, each of which can assume a binary value. This is due to the fact that the images show handwritten numbers in black on a white background. \cite{deng_mnist_2012}

As the implementation here is in PyTorch, the "Large-scale CelebFaces Attributes" (CelebA) dataset is then used to enable the generation of faces. This is easy to replace in the development environment \cite{torch_contributors_datasets_nodate}. 200,000 pictures of celebrities are available here. Similar to the FFHQ dataset, the accessories, hairstyles, ethnicity, facial expressions, etc. are also varied in the CelebA dataset. Additional annotations are included. Karras et al. also mention this dataset in their paper, noting that the images they collected in the FFHQ dataset contain higher variation. \cite{karras_style-based_2018, liu_large-scale_2015}

\section{Structure and training of a StyleGAN}
\label{sec: Structure and training of a StyleGAN}
The first part of the project deals with the training of a StyleGAN in order to be able to examine the generator in more detail. The aim is to generate images that are as realistic as possible so that appropriate weights are available that can be analyzed.

\subsection{Implementation in the Python programming language}
\label{subsec: Implementation in the Python programming language}
In a first approach, the Cheong's StyleGAN, which is implemented in the Python programming language, is examined. This implementation can be found in the GitHub repository associated with the book under the name “Faster StyleGAN” (see \cite{cheong_hands--image-generation--tensorflow-20_2023}), as some improvements in terms of structure and training time have already been implemented here. In the original code, Cheong uses the CelebA-HQ dataset described in \autoref{sec: Description of the datasets used}. This is replaced by the FFHQ dataset in the appropriate places in order to test whether the dataset developed by NVIDIA achieves better images. \cite{cheong_hands-image_2020}

The library versions used in the implementation can be found in the README file of \cite{cheong_hands--image-generation--tensorflow-20_2023}. However, since the use of TensorFlow 2.2 leads to compatibility problems with the Python version 3.11 used under Ubuntu, the program will be revised. This means that functions that are no longer available will have to be replaced by new ones. For example, an error message occurred during the execution of the method \lstinline|tf.tile|. For this reason, a class is created here that takes over this task.

\begin{figure}[htbp]
	\centering
	\includegraphics[width=0.48\textwidth]{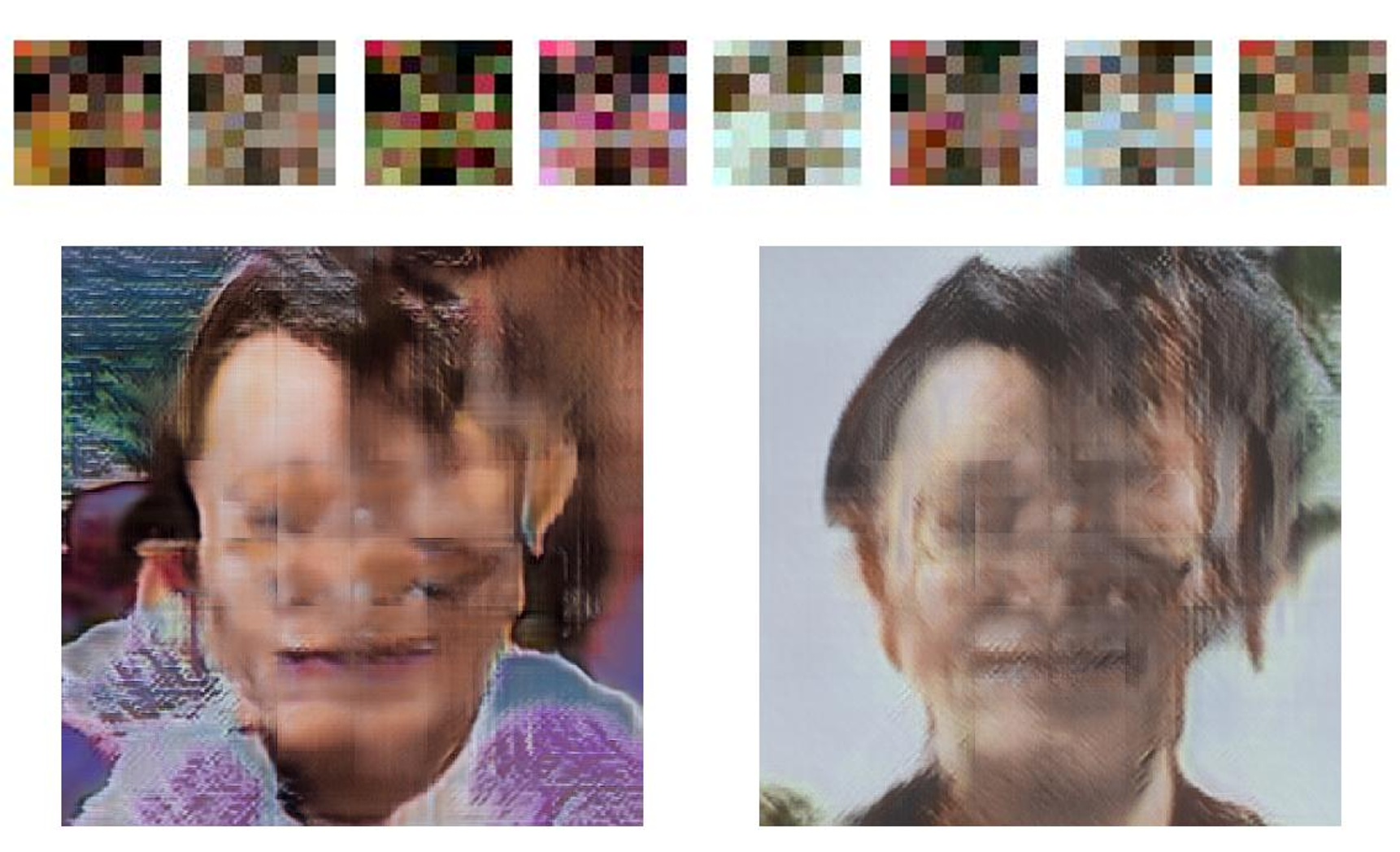}
	\caption{Faces with a resolution of $ 8 \times 8 $ (top) and $ 512 \times 512 $ (bottom), which were generated with the customized StyleGAN from Cheong, in comparison}
	\label{fig: Cheong-Generated_images}
\end{figure}

The training is progressive, which means that images are initially generated at a low resolution of $ 8 \times 8 $, and finally at a resolution of $ 512 \times 512 $, as shown in \autoref{fig: Cheong-Generated_images}. \cite{cheong_hands-image_2020}

As the resulting images obtained with this network are not satisfactory, the StyleGAN developed by NVIDIA is used in the next step. This generates very realistic images with the FFHQ dataset, as demonstrated by \cite{karras_style-based_2018}. The necessary Python files, pre-trained networks, etc. are downloaded for this purpose from the Git repository (including a reference to a Google Drive folder). The system requirements can also be found here, which are implemented accordingly in a virtual environment. Essentially, the requested Python version 3.6 and TensorFlow version 1.10.0 are used to prevent possible errors. \cite{karras_nvlabsffhq-dataset_2019}

However, it turns out that this implementation is also not compatible with the 64-bit operating system Ubuntu 22.04.5 LTS and is therefore not executable. Another possible cause is that the Cuda version V11.5.119 does not harmonize with the recommended TensorFlow version 1.10. This is confirmed by the table that can be found under \cite{tensorflow_build_2024}. As replacing instructions may again lead to poorer quality results, an alternative solution for generating images is described in the following section.

\subsection{Generation of images using a PyTorch implementation}
\label{subsec: Generation of images using a PyTorch implementation}
As the implementations in Python are proving difficult to realize in the existing environment, an alternative in PyTorch is being sought. PyTorch also uses Python syntax and can work with numerous Python libraries. It is often used in the context of machine learning, as it is very similar to TensorFlow. Tensors, multidimensional arrangements of numbers, are used as the basic unit for calculations, which are particularly suitable for representing RGB images. The excellent GPU support in particular also proves useful here. \cite{nvidia_pytorch_2025, bergmann_was_2023}

During the search, a StyleGAN is found in the targeted framework, which can be viewed at \cite{nivedwho_mystyleganipynb_2021}. Another aspect for the use is that the author refers to the StyleGAN developed by NVIDIA. Its generator structure is described in detail, based on the existing code, as this implementation is the focus of the interpretation: The basic block, labeled \lstinline|G_Block|, is the central unit that first upscales the input with a modulated convolution layer before another modified convolution further transforms the features. Two independent noise sources add random variations. A final $ 1 \times 1 $ convolution then converts the features into image channels, while an optional bilinear upsampling stage ensures that the image resolution of the upscaled features is adjusted. The activation function is applied after each transformation, and if an additional image signal is present, it is upscaled and added to the final output image. The overall generator architecture combines several of these blocks to incrementally generate a high-resolution image from a trainable constant. A mapping network transforms a random latent vector into a style representation that is used for modulation in each layer. Style mixing allows multiple latent codes to be combined for greater image variety, while the truncation trick controls style variation (see \cite{karras_style-based_2018}). Additionally, a second generator copy is used to store stabilized weights for more consistent image quality.

Nevertheless, numerous adjustments must be made to enable the generation of faces:
\begin{itemize}
	\item Since the original network works with the MNIST dataset, this must be replaced by a dataset with images of faces. The decision was made to use the CelebA dataset instead of the FFHQ dataset, as this is included in the torchvision package \cite{torch_contributors_datasets_nodate}. This can be downloaded in advance from the Google Drive folder linked under \cite{liu_large-scale_2015}, whereby \lstinline|download=False| must then be set. It is important that the TXT files with the annotations are also stored in order to ensure that \lstinline|dataset = CelebA('data', transform=transform, download=False)| works.
	\item As these are not grayscale images but RGB images, the value of \lstinline|img_channels| must be increased from 1 to 3.
	\item In addition, three mean values and standard deviations are required for RGB images, as there are three channels in contrast to the MNIST dataset. For this reason, these values are replaced by \lstinline|mean = [0.485, 0.456, 0.406]| and \lstinline|std = [0.229, 0.224, 0.225]| as can be found in several forum posts (cf. e.g. \cite{stack_exchange_inc_why_2019}).
	\item The image size is also adjusted and cropped to $ 128 \times 128 $. The transformation is outsourced to a separate function for this purpose:
	
	\lstinline|	transform = T.Compose([|\\
	\lstinline|		T.Resize(128),|\\
	\lstinline|		T.CenterCrop(128),|\\
	\lstinline|		T.ToTensor(),|\\
	\lstinline|		T.Normalize(mean, std)|\\
	\lstinline|	])|
	\item The previous point is due to the increase in the highest resolution \lstinline|max_res| from 28 to 128.
	\item The number of layers in the mapping unit and the size of the latent vectors are based on the original StyleGAN from NVIDIA. This results in the following values: \lstinline|n_layers = 8| and \lstinline|latent_size = 512|.
	\item \lstinline|group_size = 8| and \lstinline|batch_size = 80| are better suited to improving the performance of the training. This can be determined by testing different values several times.
	\item Due to the higher complexity of the CelebA dataset compared to the MNIST dataset, the size of the network is also increased. This is done by increasing the number of \lstinline|blocks| from 3 to 5, which is also the highest possible number, as this is achieved and limited by \lstinline|assert max_res <= min_res*2**blocks and max_res >= (min_res-1)*2**blocks|.
	\item For the same reason, the number of iterations for the training \lstinline|max_iter| in the training parameters \lstinline|train_params| is also increased from 50,000 to initially 100,000. However, these are not carried out fully, as there is always an abort without an error message at approx. 76,000 iterations, even after repeated testing. It is therefore assumed that the cause lies in excessive GPU utilization. In order to improve the images even further, the checkpoint file with the stored weights is used again and again to further improve the network through training. In each case, 25,000 iterations are carried out. Table 1 provides an overview of the course and duration of the training. 
	
	\autoref{tab: Itertaionen} summarizes the training process. This shows that numerous steps and - despite the GPU - approx. 56.5 hours are required to achieve improvements compared to the previous program runs.
	
	\begin{table}[htbp]
		\caption{Overview of the training process}
		\begin{center}
			\begin{tabular}{|c|c|}
				\hline
				\textbf{Number of iterations}&\textbf{Duration} \\
				\hline
				76,104 &  28\,h 30\,min 23\,s \\
				\hline
				25,000 & 9\,h 20\,min 25\,s \\
				\hline
				25,000 & 9\,h 20\,min 23\,s \\
				\hline
				25,000 & 9\,h 20\,min 6\,s \\
				\hline\hline
				\textbf{151,104} & \textbf{56\,h 31\,min 17\,s} \\
				\hline
			\end{tabular}
			\label{tab: Itertaionen}
		\end{center}
	\end{table}
	\item When retraining, the part that reinitializes the weights is also deleted to ensure that the weights already trained are used.
\end{itemize}

\section{Interpretation of the StyleGAN}
\label{sec: Interpretation of the StyleGAN}
Once suitable images have been generated and selected, we can move on to the main part of the work, the interpretation. For this purpose, the latent vector and the weights are examined in more detail in the following.

\subsection{Analysis and variation of the latent vector}
\label{subsec: Analysis and variation of the latent vector}
\begin{figure}[htbp]
	\centering
	\includegraphics[width=0.48\textwidth]{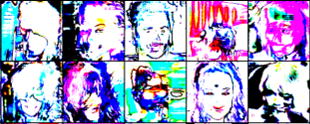}
	\caption{Faces with incorrect coloration}
	\label{fig: latent_unscaled_weights}
\end{figure}

In order to examine the latent vector in more detail, the structure of the StyleGAN, configurations and the trained weights must first be loaded. Then 32 latent vectors are generated by calling the provided function \lstinline|G.Src_Net.sample_dlatents(32)|. The \lstinline|to_img(G.Src_Net.generate(latent_vectors))| instruction is then used to output the images that generate these original latent vectors. Here, however, it turns out that the coloring is incorrect and only contours of the faces are recognizable, as shown in \autoref{fig: latent_unscaled_weights}.

The result indicates incorrect scaling of the values. This can be remedied by applying the so-called “Equalized Learning Rate” function, which is described in \autoref{subsec: The Equalized Learning Rate}, to the weights of the generator and in principle also to those of the discriminator before applying the StyleGAN.

With this change, the influence of the latent vector can now be examined more closely by multiplying it as a whole with different scaling factors. These are defined with \lstinline|scaling_factors = [0.05, 0.10, 0.25, 0.5, 1.0, 1.5, 2.5, 5.0, 10.0]|, then run through in a \lstinline|for| loop and the images are output.

In addition, the individual dimensions of the vector are examined in more detail and what the change in these dimensions changes on the image. An interactive visualization is set up for the purpose of controlling a dimension. "Jupyter Widgets" is used for this by importing the \lstinline|ipywidgets| library, which provides sliders and checkboxes, for example. \cite{project_jupyter_jupyter_nodate}

\begin{figure}[htbp]
	\centering
	\includegraphics[width=0.48\textwidth]{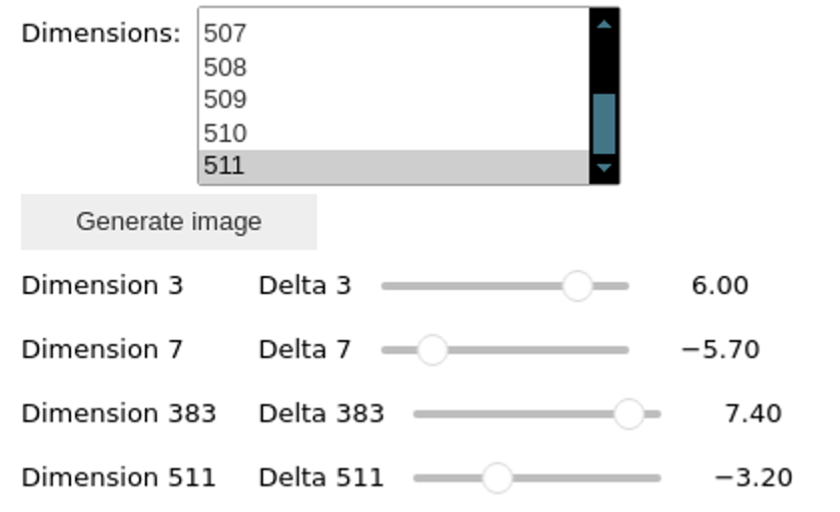}
	\caption{Interactive control of one or more dimensions of the latent vector using the \lstinline|SelectMultiple()| and \lstinline|FloatSlider()| functions, among others}
	\label{fig: interactive_visualization}
\end{figure}

\autoref{fig: interactive_visualization} shows the interactive control. One or more dimensions of the latent vector to be modified can be selected. Sliders are then output for these, with which values between -10 and +10 can be added in steps of 0.1. By adjusting the command \lstinline|delta_slider = widgets.FloatSlider(value = 0.0, min = -10.0, max = 10.0, step = 0.1, description = f'Delta {dim}')|, the minimum and maximum values and the step width of the slider can also be easily adjusted for test purposes.

Both the original image and the modified image are output when the “Generate image” button is clicked. The vector is also displayed for verification purposes. In this way, differences can be clearly identified and assigned to the appropriate dimension.

\subsection{Examination of the weights of the generator}
\label{subsec: Examination of the weights of the generator}
Similar to the investigation of the latent vector, precautions must also be taken here, such as loading the network. In order to be able to examine the weights of the generator in more detail, their distribution is first considered. The checkpoint file is used for this purpose. The weights of both the generator and the discriminator are regularly stored in this file during training, which has also made it possible to continue training the network. This must first be examined more closely so that the corresponding weights can be loaded. The layer keys from the state dictionary are output first. For the last block, a section looks like this: \\
\lstinline|Src_Net.4.upconv.bias| \\
\lstinline|Src_Net.4.upconv.weight_orig| \\
\lstinline|Src_Net.4.upconv.style.bias| \\
\lstinline|Src_Net.4.upconv.style.weight_orig| \\
\lstinline|Src_Net.4.conv.bias| \\
\lstinline|Src_Net.4.conv.weight_orig| \\
\lstinline|Src_Net.4.conv.style.bias| \\
\lstinline|Src_Net.4.conv.style.weight_orig| \\
\lstinline|Src_Net.4.noise.noise_strength| \\
\lstinline|Src_Net.4.noise2.noise_strength| \\
\lstinline|Src_Net.4.to_channels.bias| \\
\lstinline|Src_Net.4.to_channels.weight_orig| \\
\lstinline|Src_Net.4.to_channels.style.bias| \\
\lstinline|Src_Net.4.to_channels.style.weight_orig| \\
This naming of the weights will also play a role in pruning later on. For now, the distribution of the weights will be analysed. To do this, histograms are created and certain values, such as the minimum and maximum values, the mean value and the number of weights, are output.

\begin{figure}[htbp]
	\centering
	\includegraphics[width=0.48\textwidth]{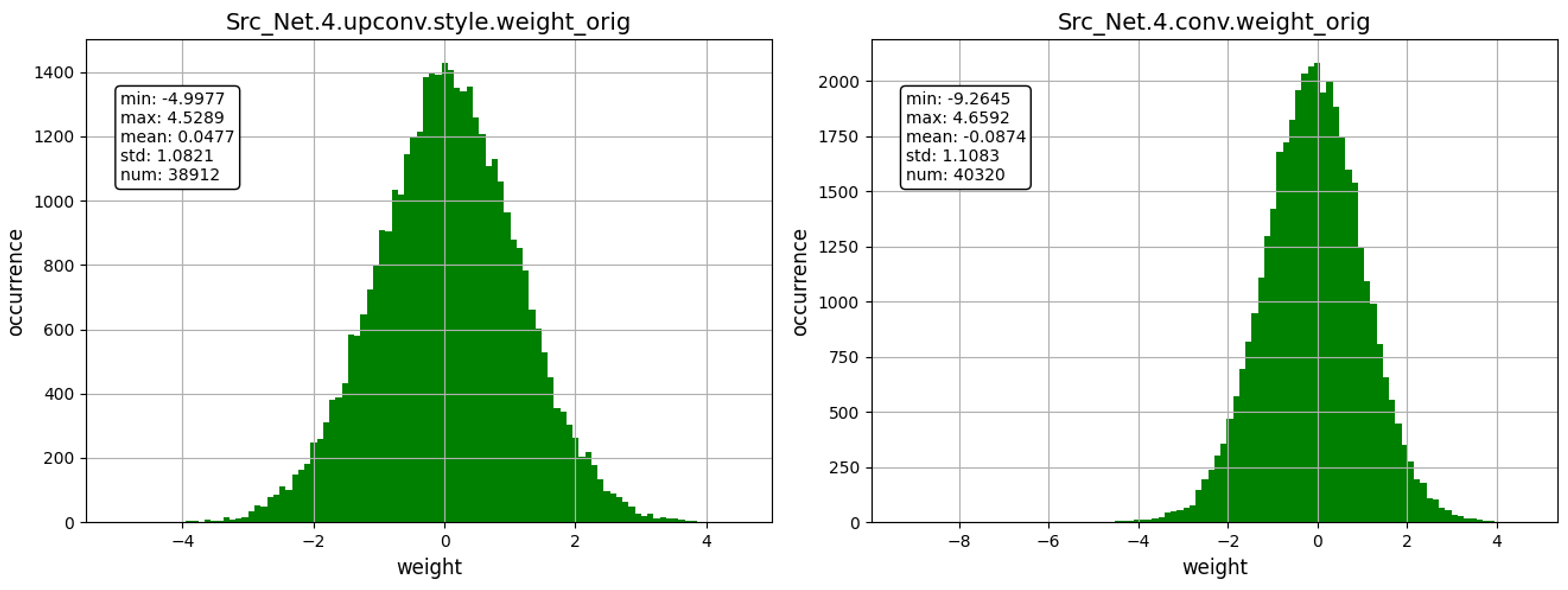}
	\caption{Exemplary representation of the distribution of weights for two layers of the last block}
	\label{fig: weights_distribution}
\end{figure}

It is striking that the mean value is approx. 0 and the standard deviation approx. 1 for almost all layers of the last block, for example. However, this ensures that pruning can be carried out as described in \autoref{subsec: Pruning for the reduction of computing power}. The illustration makes it clear that there are enough weights close to zero that can possibly be neglected.

As already mentioned, there are problems with the naming convention of the stored weights: these are all named \lstinline|weight_orig| due to the application of the Equalized Learning Rate function. This leads to an error message called \lstinline|KeyError|. The attempts to rename the keys also failed, which is why the pruning method itself is implemented, as the PyTorch library \lstinline|prune| cannot be used with these names. This approach is based on the procedure described in \cite{paganini_pruning_2023}, for example.

The \lstinline|prune_generator| function expects two transfer parameters: the model, in other words the generator, and a limit value. This determines the absolute value up to which the values are set to zero. A mask containing the corresponding binary values is then created for each module. If the weights are then multiplied by this mask, only those remain that are greater in amount than the set limit value: \\
\lstinline|mask = (module.weight_orig.abs() >= threshold).float()| \\
\lstinline|module.weight_orig *= mask|.\\
Furthermore, a function is defined that finally counts the weights that are not equal to zero. This can later be used to observe how the size of the network decreases. A fixed latent vector is then defined to ensure that the pruning can always be observed using the same images. The limit values 0 to 1.0 are then run through in 0.001 steps in a \lstinline|for| loop and the generator is pruned. To make it easier to verify the results, the output of the discriminator is saved in addition to the limit values and the number of remaining weights. As there are 32 images in each case, the mean value is calculated. To obtain the actual probabilities, the sigmoid function must be applied, which can be taken from the reference code: \\
\lstinline|logits = D.get_score(G_copy.Src_Net.generate(fixed_latents))| \\
\lstinline|scores = torch.sigmoid(logits)|. \\
After every 20 limit values, the corresponding images are still output. Finally, two plots are displayed which show the mean value of the discriminator result and the number of weights still present above the limit value over the course.

\section{Results}
\label{sec: Results}
This section discusses the methods used in the previous chapter as well as the results obtained and their significance.

\subsection{Evaluation of the generated images}
\label{subsec: Evaluation of the generated images}
\begin{figure}[htbp]
	\centering
	\includegraphics[width=0.48\textwidth]{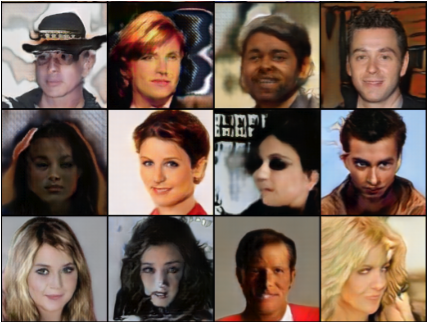}
	\caption{Excerpt from the images generated with StyleGAN and PyTorch}
	\label{fig: Generated images}
\end{figure}

The method described in \autoref{subsec: Generation of images using a PyTorch implementation} generates images that are clearly identifiable as artificial as well as images that are very close to reality. \autoref{fig: Generated images} shows pictures for both cases.

The images in the bottom left and top right corners, for example, can hardly be distinguished from real faces. If necessary, the background of the man, which is partially blurred with his hair, can be seen as an argument for a generated image. On the other hand, the image in the second row and third column, for example, is clearly recognizable as generated. The face and facial features are distorted. The blurred background is also an indication of an artificially created face. The first image in the first row also shows an attempt to create glasses as an accessory. However, this is only very faintly recognizable and is only hinted at.

\begin{figure}[htbp]
	\centering
	\includegraphics[width=0.35\textwidth]{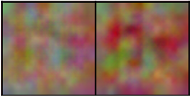}
	\caption{Images at the beginning of the training with random color values of the individual pixels}
	\label{fig: Generated images begin}
\end{figure}

To show how the training has developed, \autoref{fig: Generated images begin} shows the images that are generated at the beginning of the training.

\begin{figure}[htbp]
	\centering
	\includegraphics[width=0.48\textwidth]{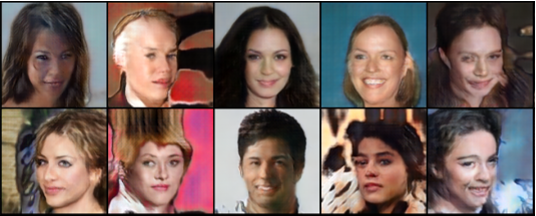}
	\caption{Images after the first training run with clearly recognizable defects in large areas}
	\label{fig: Generated images first process}
\end{figure}

As you can see, the individual pixels have been given random color values and nothing is visible. However, after the first training process with approx. 76,000 iterations, faces can already be clearly recognized (see \autoref{fig: Generated images first process}). In some cases, however, entire areas are left out and filled with blue color. This is particularly evident in the images in the top left, bottom left and bottom right corners.

\begin{figure}[htbp]
	\centering
	\includegraphics[width=0.45\textwidth]{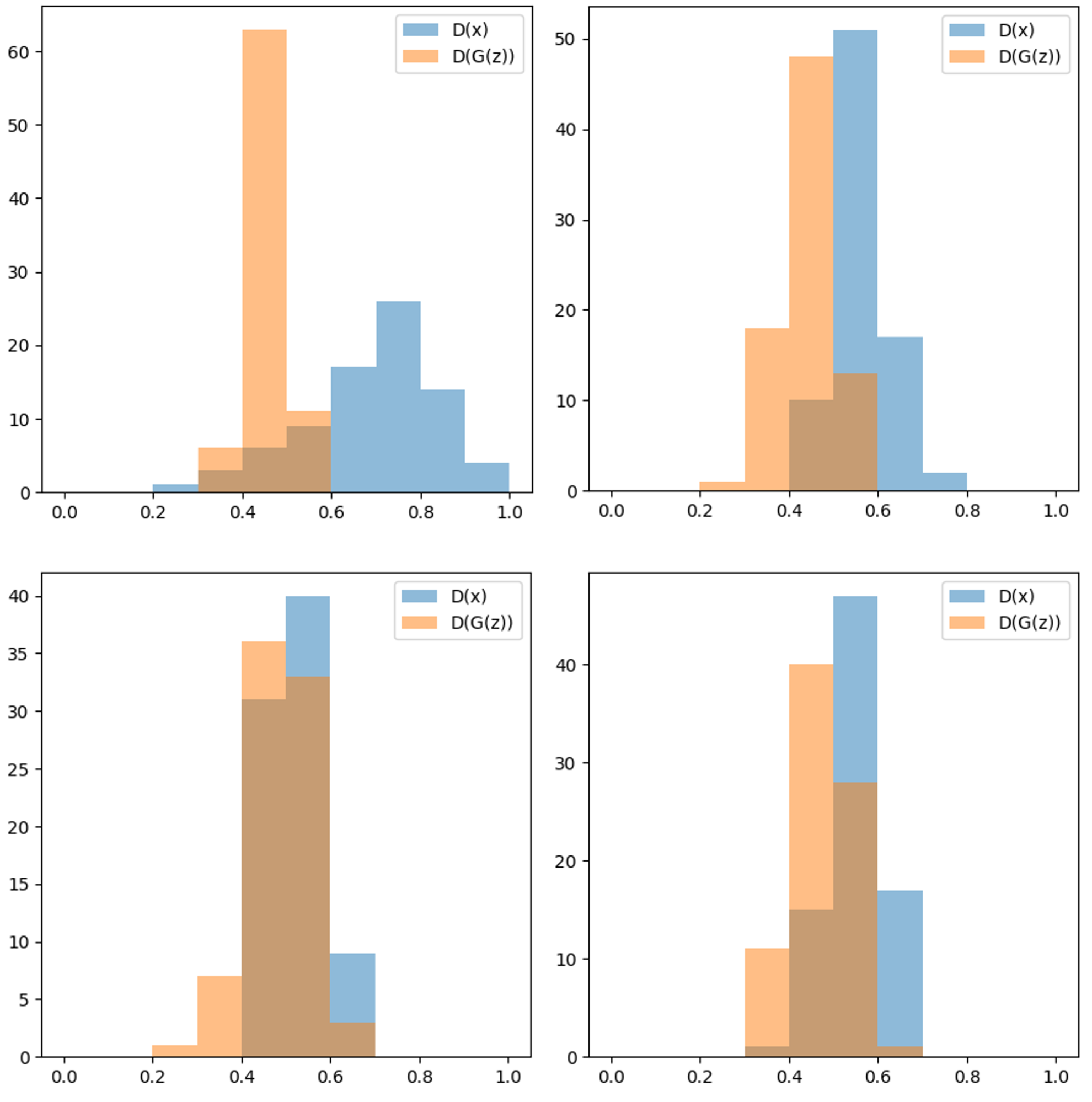}
	\caption{Comparison of the discriminator results (top left: at the start of training, top right: after the first training process with approx. 76,000 iterations, bottom left:  after the second training process and a total of approx. 100,000 iterations, bottom right: after the end of training)}
	\label{fig: Results D}
\end{figure}

In addition, the notebook used contains the function that the results of the evaluations of $ D(\mathbf{x}) $ and $ D(G(\mathbf{z})) $ are compared. To illustrate the improvement resulting from the training, these are also shown graphically for the three different stages in \autoref{fig: Results D}. The discriminator can be used as an indicator or “signal” of how well the generator has been trained. \cite{prince_understanding_2024}

In the top left image at the beginning of training, the discriminator can distinguish between real and generated images well, so that the values of $ D(\mathbf{x}) $ (blue) are usually high, while $ D(G(\mathbf{z})) $ (orange) are low. This means that the discriminator recognizes real images as “real” and classifies generated images as “false”. In the second upper diagram, the distributions of $ D(\mathbf{x}) $ and $ D(G(\mathbf{z})) $ converge and overlap slightly. It can be concluded that the generator improves and the generated images become more convincing. In contrast, the discriminator becomes less reliable. The image in the lower left corner shows an almost complete overlap of the two distributions, which indicates a good equilibrium state of the StyleGAN. The bottom right picture shows that the distributions are moving away from each other again. This may be due to overfitting. This can be combated by paying attention to the hyperparameters. For example, it is possible to adjust the learning rate at this stage of training. \cite{tomczak_deep_2024}

\begin{figure}[htbp]
	\centering
	\includegraphics[width=0.48\textwidth]{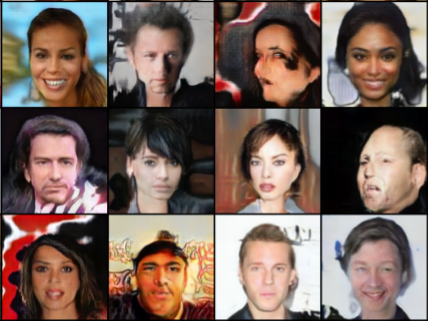}
	\caption{Images for interpretation from the second stage of training}
	\label{fig: Generated images for interpretation}
\end{figure}

For this reason, the images taken after the second training run are used for interpretation. A section of these is shown in \autoref{fig: Generated images for interpretation}. This selection of images obviously contains faulty areas (see picture in the first row and third column), but this cannot be ruled out even in the version after the end of training.

\subsection{Effects of pruning}
\label{subsec: Effects of pruning}
\begin{figure}[htbp]
	\centering
	\includegraphics[width=0.45\textwidth]{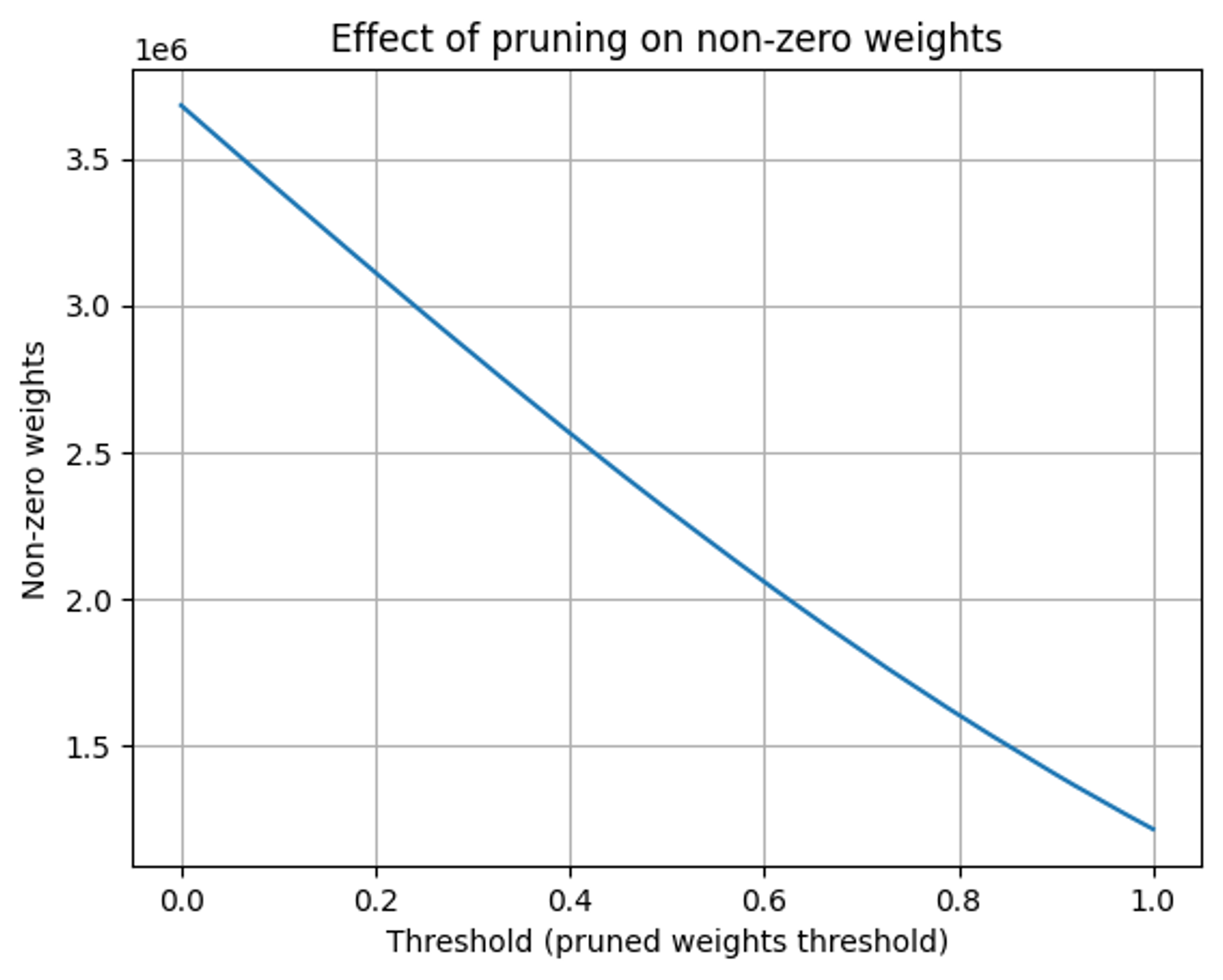}
	\caption{Progression of the number of weights that are not overwritten with zero above the limit value used for pruning (at the beginning: 3,680,500 non-zero weights, at the end: 1,218,002 non-zero weights)}
	\label{fig: plot_non-zero_weights}
\end{figure}

Initially, the generator works with 3,680,500 weights, which means that no limit value is applied yet. If you look at the progression in \autoref{fig: plot_non-zero_weights}, you can see that the number of weights decreases almost linearly as the limit value increases. At the end, with a limit value of 1.0, 1,218,002 weights are still available, which corresponds to slightly more than a third of the original weight parameters.

\begin{figure}[htbp]
	\centering
	\includegraphics[width=0.45\textwidth]{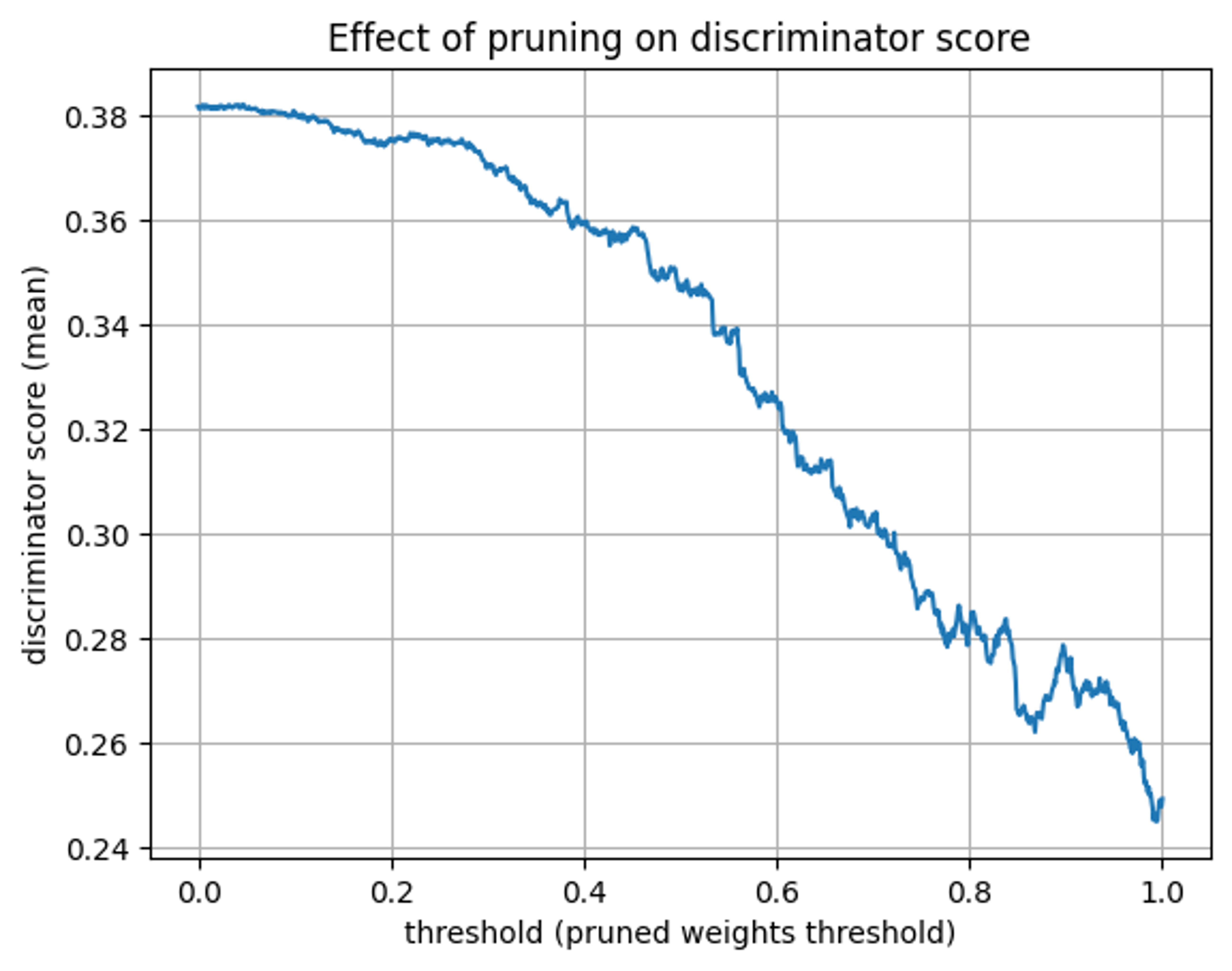}
	\caption{Development of the mean value of the discriminator result for the 32 generated images above the limit value for pruning (at the beginning: 38.18\,\%, at the end: 24.93\,\%)}
	\label{fig: plot_D_score}
\end{figure}

If you now look at the course of the discriminator result in \autoref{fig: plot_D_score}, the effects of this measure can be clearly seen.

\begin{figure}[htbp]
	\centering
	\includegraphics[width=0.4\textwidth]{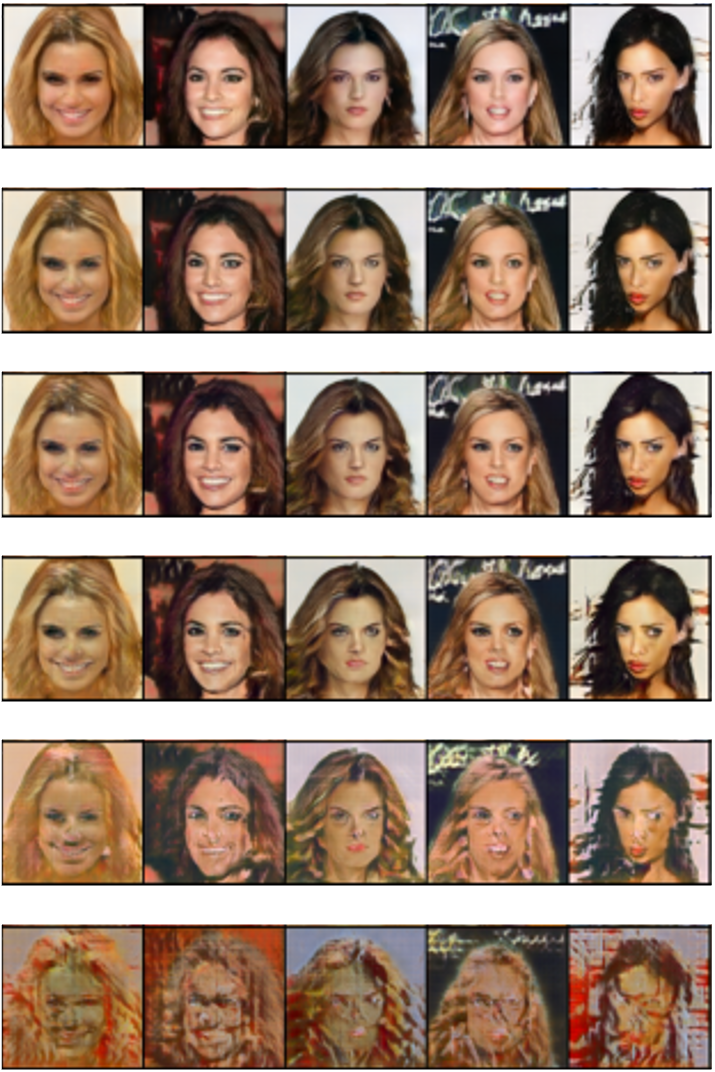}
	\caption{Excerpts from each of the 32 images to illustrate the effect of pruning (limit values: row 1: 0.0 (no limit value), row 2: 0.3, row 3: 0.4, row 4: 0.5, row 5: 0.7, row 6: 1.0)}
	\label{fig: effect_of_pruning}
\end{figure}

Up to a limit value of approx. 0.4, the results are still acceptable. The probability here is 35.98\,\%. After that, a drop can be seen, indicating that the discriminator is finding it increasingly easy to recognize the fake images as generated. \autoref{fig: effect_of_pruning} also shows some pictures at this point to verify the results, always using the same section of the 32 images.

In the first row, with a pruning limit of 0.0, the images appear highly realistic. The facial features are sharp and well-defined, and the lighting and backgrounds are natural. This row represents the unpruned model, where all weights are intact, allowing for the generation of highly detailed and convincing human faces.

In the second row, with a pruning limit of 0.3, the images still look realistic, but some minor loss in detail becomes noticeable. Facial expressions are slightly less refined, and certain textures, such as hair or skin, may appear a bit smoother. Nonetheless, the overall realism remains high, and the images are still clearly recognizable as human faces.

As you move to the third and fourth rows, corresponding to pruning limits of 0.4 and 0.5 respectively, the degradation becomes more evident. Facial features begin to lose structure and symmetry, and artifacts start to appear. Although the general shape of the faces is preserved, the images look less natural, with noticeable distortions in the eyes, mouth, or hair.

In the fifth row, with a pruning limit of 0.7, the images become significantly distorted. While some basic elements of a face are still visible, the realism is greatly reduced. The model struggles to generate coherent features, resulting in uncanny or surreal representations that clearly diverge from human-like appearances.

Finally, in the sixth row with a limit value of 1.0, the images are heavily corrupted. The faces are barely recognizable, with intense artifacts, color bleeding, and structural collapse. The output no longer resembles real people and appears abstract.

Overall, this sequence demonstrates that pruning - while useful for reducing model complexity - has a strong impact on the quality of generated images. As the pruning limit increases, the model loses critical information needed for fine detail and coherence, ultimately compromising the realism of the generated faces. This evaluation thus confirms the results of the discriminator in \autoref{fig: plot_D_score}.

\subsection{Scaling of the entire latent vector}
\label{subsec: Scaling of the entire latent vector}
\begin{figure}[htbp]
	\centering
	\includegraphics[width=0.48\textwidth]{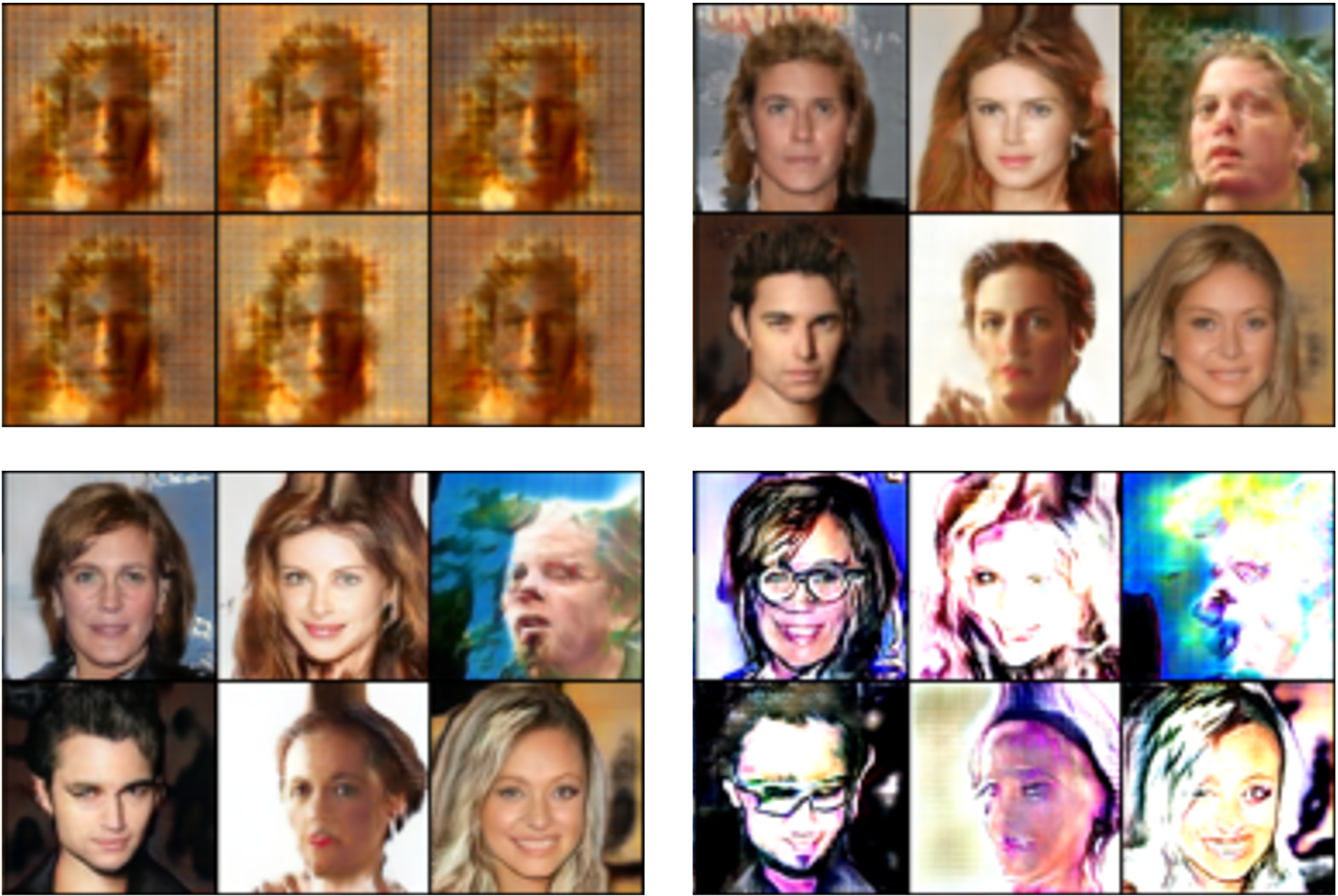}
	\caption{Scaling of the latent vector with the factor 0.05 (top left), 0.5 (top right), 1.0 or no scaling (bottom left) and 5.0 (bottom right)}
	\label{fig: scale_whole_latent}
\end{figure}

The resulting images are considered here, which are created when the latent vector is multiplied by a scaling factor. \autoref{fig: scale_whole_latent} shows some excerpts. It should be noted that the same examples from the 32 generated images are always shown.

This graphic representation illustrates how scaling the latent vector affects the output of a generative model. Starting in the top left, where the latent vector is scaled by a factor of 0.05, the generated images appear extremely blurry and vague. Facial features are barely recognizable, and the outputs resemble a diffuse average of many faces rather than any coherent individual. As the scale increases to 0.5, the faces begin to form more clearly, but they still lack sharpness and fine detail, appearing soft and somewhat flattened.

In the bottom left, where no scaling is applied, the faces are sharp, realistic, and rich in detail. Facial structure, texture, and expression are clearly visible. This quadrant serves as a reference for what the model produces under normal conditions.

When the latent vector is strongly amplified witch a scaling factor of 5.0, the outputs become increasingly stylized and distorted. Faces start to break apart, with unnaturally vivid colors and exaggerated contrasts. Notably, eyeglasses become much more prominent and stylized - almost cartoon-like or abstract - which suggests that the generator heavily overemphasizes certain visual cues under high latent magnitudes. This exaggeration affects not only accessories like glasses but also facial contours and lighting, leading to surreal or painterly renderings.

In summary, low latent vector magnitudes suppress meaningful features, leading to blurry outputs, while high magnitudes amplify and distort them, sometimes producing creative but unrealistic artifacts such as overemphasized glasses and exaggerated facial shapes.

\subsection{Changing individual dimensions of the latent vector}
\label{subsec: Changing individual dimensions of the latent vector}
Since there are countless ways to change the latent vector in one or more dimensions, this section is limited to looking at the effect of individual components. This can be done with the interactive control. It is important here that the counting starts with zero, so that the highest dimension is 511:
\begin{itemize}
	\item \textbf{Dimension 3:} In many images, increasing this dimension has an effect on the hair. As a result, some generated persons have less hair or a high forehead, as shown in \autoref{fig: dim3}.
	\begin{figure}[htbp]
		\centering
		\includegraphics[width=0.4\textwidth]{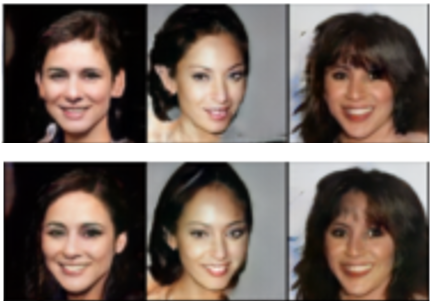}
		\caption{Increase in dimension 3 by 10.0 (above: before the change, below: after the change)}
		\label{fig: dim3}
	\end{figure}
	\item \textbf{Dimension 70:} Changing this dimension causes the image to become brighter or darker, as can be seen in \autoref{fig: dim70}. This also refers to the glasses, some of which are added and some of which disappear again.
	\begin{figure}[htbp]
		\centering
		\includegraphics[width=0.4\textwidth]{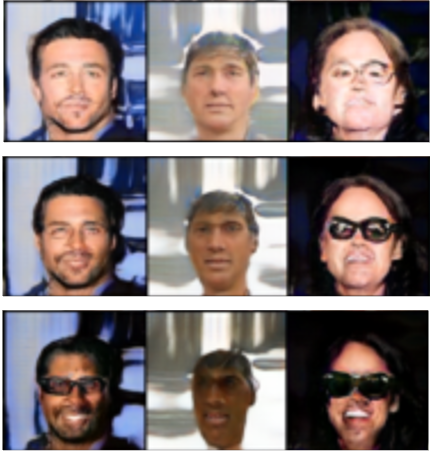}
		\caption{Change of dimension in both directions (top: subtraction of 10, center: original image: bottom: addition of 10)}
		\label{fig: dim70}
	\end{figure}
	\item \textbf{Dimension 99:} This can reduce red spots in the background and make the display more realistic. (see \autoref{fig: dim99})
	\begin{figure}[htbp]
		\centering
		\includegraphics[width=0.3\textwidth]{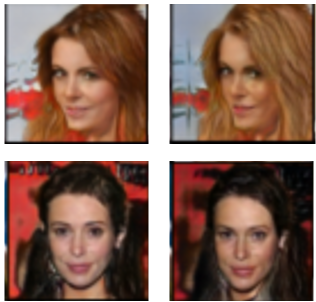}
		\caption{Reduction of defects by increasing the dimension by 10 (left: before, right: after)}
		\label{fig: dim99}
	\end{figure}
	\item \textbf{Dimension 125:} Reducing this dimension results in facial hair or a beard. In some images where a beard is present, it becomes darker, as shown in \autoref{fig: dim125}.
	\begin{figure}[htbp]
		\centering
		\includegraphics[width=0.4\textwidth]{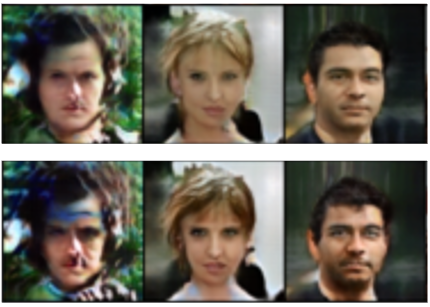}
		\caption{Change in facial hair (top: original image, bottom: after reducing the dimension by 10)}
		\label{fig: dim125}
	\end{figure}
	\item \textbf{Dimension 360:} By increasing the dimension, the images are given a stronger contrast. In some cases, this makes them appear even more realistic. However, this does not compensate for imperfections. (see \autoref{fig: dim360})
	\begin{figure}[htbp]
		\centering
		\includegraphics[width=0.4\textwidth]{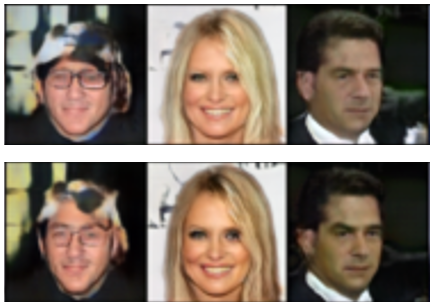}
		\caption{Contrast adjustability by this dimension (top: original image, bottom: after increasing the dimension by 10)}
		\label{fig: dim360}
	\end{figure}
\end{itemize}

To summarize at this point, it can be said that changing individual dimensions makes it possible to vary the desired accessories, hair, etc. In addition, various effects such as exposure or contrast can be adjusted.

\begin{figure}[htbp]
	\centering
	\includegraphics[width=0.35\textwidth]{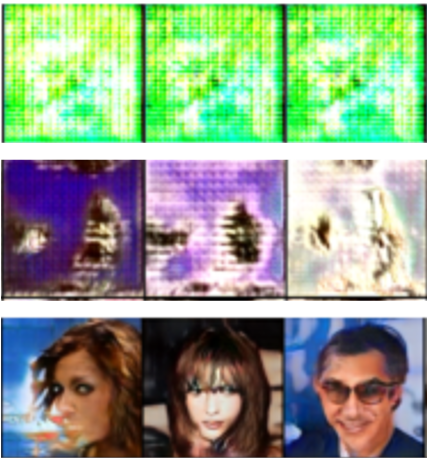}
	\caption{Test of different deltas to find a limit value (top: addition by 1000, middle: addition by 100, bottom: addition by 20)}
	\label{fig: deltas}
\end{figure}

Finally, the extent to which the dimensions can be increased or reduced is examined. It turns out that with an increase of 1,000, faces are no longer recognizable. The color changes depending on the dimension, as can be seen in \autoref{fig: deltas}. Even a change of 100 does not result in any recognizable faces. With a change of 20, people are recognizable, but in significantly poorer quality. For this reason, the interval used is a good choice for examining the latent vector.

\section{Conclusion}
\label{sec: Conclusion}
When it comes to generating images, it can be said that once a suitable network has been found or implemented, this is no longer a major effort. The only problem is finding suitable hyperparameters and having enough working memory available for training. With the knowledge that can be gained from the inter-share control system, you can also build your own tools for targeted manipulation. Deepfakes, created using AI or also a StyleGAN, pose serious risks. They can be used in misinformation campaigns to manipulate public opinion - for example, by making political figures appear to say false things. They also enable realistic identity fraud, such as impersonation in video calls or fake audio recordings, undermining trust in digital communication. After closer examination of a StyleGAN, desired changes and faces can be generated, which is a danger in many cybersecurity scenarios. \cite{medium_generative_2024}

\phantomsection
\addcontentsline{toc}{section}{References}
\printbibliography

\end{document}